\documentclass{IOS-Book-Article}

\usepackage{mathptmx}
\usepackage{soul}\setuldepth{article}

\usepackage{cite}

\usepackage{bm}
\usepackage{graphicx}
\usepackage{subcaption}

\usepackage{multicol}
\usepackage{hyperref}

\usepackage[hashEnumerators,smartEllipses]{markdown}
\usepackage{amssymb}
\def\hb{\hbox to 11.5 cm{}}

\begin{document}

\pagestyle{headings}
\def\thepage{}

\begin{frontmatter}              % The preamble begins here.

%\pretitle{Pretitle}
% \title{Instructions for the Preparation of an\\
% Electronic Camera-Ready Manuscript in\\ \LaTeX}
\title{A Machine With Human-Like Memory Systems}

\markboth{}{April 2022\hb}
%\subtitle{Subtitle}

\author[A]{\fnms{Taewoon} \snm{Kim}},
\author[A]{\fnms{Michael} \snm{Cochez}},
\author[A]{\fnms{Vincent} \snm{François-Lavet}},
\author[B]{\fnms{Mark} \snm{Neerincx}}, and
\author[A]{\fnms{Piek} \snm{Vossen}}

% \author[A]{\fnms{Book Production} \snm{Manager}%
% \thanks{Corresponding Author: Book Production Manager, IOS Press, Nieuwe Hemweg 6B,
% 1013 BG Amsterdam, The Netherlands; E-mail:
% bookproduction@iospress.nl.}},

% \author[B]{\fnms{Second} \snm{Author}}
% and
% \author[B]{\fnms{Third} \snm{Author}}

% \runningauthor{B.P. Manager et al.}
% \address[A]{Book Department, IOS Press, The Netherlands}
\address[A]{Vrije Universiteit Amsterdam}
\address[B]{Technische Universiteit Delft}
\{\texttt{t.kim, m.cochez, vincent.francoislavet, p.t.j.m.vossen}\}\texttt{@vu.nl}
\texttt{m.a.neerincx@tudelft.nl}

% \address[B]{Short Affiliation of Second Author and Third Author}

% \begin{abstract}
% These instructions are designed for the Preparation of an Electronic
% Camera-Ready Manuscript in \LaTeX{} and should be read carefully. If you
% have any questions regarding the instructions, please contact the Book
% Department, by \mbox{e-mail}: \textit{bookproduction@iospress.nl}.
% \end{abstract}

\begin{abstract}
Inspired by the cognitive science theory, we explicitly model an agent with both semantic and episodic memory systems, and show that it is better than having just one of the two memory systems. In order to show this, we have designed and released our own challenging environment, “the Room”, compatible with OpenAI Gym, where an agent has to properly learn how to encode, store, and retrieve memories to maximize its rewards. The Room environment allows for a hybrid intelligence setup where machines and humans can collaborate. We show that two agents collaborating with each other results in better performance than one agent acting alone. \textbf{The code is open-sourced at \href{https://github.com/humemai/agent-room-env-v0}{https://github.com/humemai/human-like-memory-systems}.}
\end{abstract}

\end{frontmatter}
\markboth{April 2022\hb}{April 2022\hb}

\section{Introduction}
\label{sec:intro}

In cognitive science, it is thought that humans have an explicit memory system, which is composed of semantic and episodic memory systems. Semantic memory has to do with general world knowledge, while episodic has to do with one's personal memory. For example, when one asks you a question, “In general, where are laptops located?” you might be able to answer it, if you have successfully encoded and stored a relevant memory in your brain. Let's assume that you have, and your answer is “On the desks”. However, it is likely that you do not know when and where you have encoded and stored the memory. Nonetheless, you were able to retrieve it. This is because this type of memory is semantic. When your brain deals with such factual (general) knowledge memories, it does not store the information regarding when and where. Let's ask you another question, “Where is Karen's laptop?” Let's again assume that you have observed where Karen's laptop was. To answer this question, one revisits when and where this memory was encoded and stored. Retrieval of such a memory is a reconstruction process of it. This type of memory is called episodic. It is more personal to you than semantic~\cite{Tulving1983-TULEOE, tulving1985memory, Tulving1973EncodingSA}.

% Modern machines are good at answering factual knowledge questions. Any search engines or virtual assistants can now retrieve relevant information from the Internet and answer your questions related to facts and commonsense (semantic). However, if you ask your agent if it remembers what it did yesterday (episodic), it will not give you a good answer. This is not because they are dumb, but because they were never designed to answer such questions. However, we also need machines capable of such things.

% Consider for example a robot deployed to a nursery home, and its job is to remember and remind elderly and disabled taking pills every day. If the robot observing the environment can successfully remember when and where it has seen them taking the pills, it will be a great help to them. However, it is not always feasible for a machine to encode and store every observation as an episodic memory, as its memory capacity is bounded. Then it can just use commonsense knowledge (semantic memory) and say “You left your pills in the cabinet”, which can actually be found in ConceptNet, an open commonsense knowledge graph~\cite{Speer_Chin_Havasi_2017}. Its perception is also limited, as it can not have an infinite number of sensors (e.g., eyes). Therefore, it would be wise for agents to work together.

Motivated by this, we have explicitly modeled an agent that has both semantic and episodic memory systems. An agent interacts with the environment and has to answer questions to maximize the rewards. Our hypothesis is that if it can successfully encode and store relevant observations in its brain as either semantic or episodic memories, then it can also answer the questions more successfully than using only one of the two memory systems.

The contributions of this paper are as follows. \textbf{(1)}~Inspired by the cognitive science theory, we explicitly model an agent with both semantic and episodic memory systems, and show that it is better than having just one memory system in our experiments. \textbf{(2)}~We designed and released our own challenging environment, compatible with OpenAI Gym~\cite{brockman2016openai}, where an agent has to properly learn how to encode, store, and retrieve memories to maximize rewards. \textbf{(3)}~We demonstrate that when an agent collaborates with another agent or human, it leads to better performance. 

% The rest of this paper is organized as follows. In Section~\ref{sec:methodology}, we propose our method, which allows an agent to learn how to encode observations, and to store and retrieve memories. In Section~\ref{sec:experimentalsetup}, we outline how we carried out our experiments and show the results in Section~\ref{sec:results}. In Section~\ref{sec:related}, we compare our work with other works, and show how ours differs from theirs. Finally, in Section~\ref{sec:conclusions}, we conclude this paper and mention some future works.

\section{Methodology}
\label{sec:methodology}

\subsection{The Room Environment}
\label{sec:roomenv}

The OpenAI-Gym-compatible Room environment is one big room with $N_{people}$ number of people who can freely move around. Each of them selects one object, among $N_{objects}$, and places it in one of the $N_{locations}$ locations. $N_{agents}$ number of agent(s) are also in this room. They can only observe one human placing an object, one at a time; $\bm{x}^{(t)}$. At the same time, they are given one question about the location of an object; $\bm{q}^{(t)}$. $\bm{x}^{(t)}$ is given as a quadruple, $(\bm{h}^{(t)}, \bm{r}^{(t)}, \bm{t}^{(t)}, t)$, For example, \texttt{<James's laptop, AtLocation, James's desk, 42>} accounts for an observation where an agent sees James placing his laptop on his desk at $t=42$. $\bm{q}^{(t)}$ is given as a double, $(\bm{h}, \bm{r})$. For example, \texttt{<Karen's cat, AtLocation>} is asking where Karen's cat is located. If the agent answers the question correctly, it gets a reward of $+1$, and if not, it gets $0$.

The reason why the observations and questions are given as RDF-triple-like format is two folds. One is that this structured format is easily readable / writable by both humans and machines. Second is that we can use existing knowledge graphs, such as ConceptNet~\cite{Speer_Chin_Havasi_2017}.

To simplify the environment, the agents themselves are not actually moving, but the room is continuously changing. There are several random factors in this environment to be considered: 

\begin{enumerate}
  \item With the chance of $p_{commonsense}$, a human places an object in a commonsense location (e.g., a laptop on a desk). The commonsense knowledge we use is from ConceptNet. With the chance of $1 - p_{commonsense}$, an object is placed at a non-commonsense random location (e.g., a laptop on the tree).
  \item With the chance of $p_{new\_location}$, a human changes object location.
  \item With the chance of $p_{new\_object}$, a human changes his/her object to another one.
  \item With the chance of $p_{switch\_person}$, two people switch their locations. This is done to mimic an agent moving around the room.
\end{enumerate}

All of the four probabilities account for the Bernoulli distributions.

\subsection{Episodic and Semantic Memory Systems}
\label{sec:episodic_semantic}

Each agent partially observes the environment (i.e., they cannot see the entire room at once, but one human at a time). This means that it should keep the history of its observations. This can be done by having memory systems. In our work, we model our agent to have a human-like explicit memory system. This means that it has both episodic and semantic memory systems. For example, at current time $t=23$, a question given by the environment is \texttt{<Karen's cat, AtLocation>}. If the agent has an episodic memory \texttt{<Karen's cat, AtLocation, Karen's office, 21>}, which it has seen two time steps ago, this episodic memory will likely be a “correct” memory to be retrieved to answer the question.

Not every observation has to be saved in the episodic memory system. Using our commonsense, we know that laptops are mostly placed on the desk. So for example, at current time $t=23$, let's say that a question \texttt{<James's laptop, AtLocation>} is given. If the agent has an episodic memory of this event, then it can use it. But if not, then it can use a commonsense knowledge \texttt{<laptop, AtLocation, desk, 10>}. This commonsense knowledge forms the semantic memory of the agent, since semantic memory has to do with the general knowledge of the world. Similar to episodic memories, semantic memories are also quadruples. However, the last element of a semantic memory is not a timestamp, but the strength of the semantic memory. For example, \texttt{<laptop, AtLocation, desk, 10>} is a stronger semantic memory than \texttt{<laptop, AtLocation, garage, 5>}, since the agent has seen laptops being placed on desk 10 times, while it has only seen them in a garage five times. Notice that semantic memories do not include the names of people, since this type of memory is not person-specific.

Both episodic and semantic memory systems, $\bm{M}_{E}$ and $\bm{M}_{S}$, respectively, are bounded in size. Since an agent can not store all of its observations, it should learn what to store and what to forget. It should also learn that some of its episodic memories can be summarized into one semantic memory. For example, if it always sees humans placing their laptops on their desks, then perhaps one semantic memory \texttt{<laptop, AtLocation, desk>} is enough to answer the related questions.

% Episodic and semantic memories can be stored as a knowledge graph, as in Figure~\ref{fig:knowledge_graph}. In this example figure, both Karen's and James's laptop can be considered as entities that are instances of the more generic entity “laptop”. Since the semantic memory alone can explain the two episodic memories, it would be wise for the agent to remove the two. It is also possible for the agent to exploit this data structure (e.g., using GNNs).

% \begin{figure}[tb]
% \centering
% \includegraphics[width=0.5\textwidth]{fig/episodic-semantic}
% \caption{An example memory state. This state has two episodic memories (orange) in $\bm{M}_{E}$ and one semantic memory (blue) in $\bm{M}_{S}$.}
% \label{fig:knowledge_graph}
% \end{figure}

\subsection{Hybrid Intelligence}
\label{sec:hybrid}

The Room environment does not restrict the number of agents. It allows for human-machine or machine-machine collaboration. For example, at current time $t=23$, a question given by the environment is \texttt{<Karen's cat, AtLocation>}. If $agent_{1}$ has an episodic memory \texttt{<Karen's cat, AtLocation, Karen's office, 21>} and $agent_{2}$ has an episodic memory \texttt{<Karen's cat, AtLocation, Karen's desk, 22>}, then it is more likely that the memory of $agent_{2}$ can answer the question more accurately since its memory is more recent than that of $agent_{1}$. Or, it is also possible for them to use commonsense knowledge that “Cats are often found on the lap of their owners.”, if the agents have encoded such a thing in their semantic memory system.

\section{Experimental Setup}
\label{sec:experimentalsetup}

\subsection{Data Collection and the Environment Hyperparameters}
\label{sec:data_collection}

% The knowledge graph used for the Room environment was collected from ConceptNet~\cite{Speer_Chin_Havasi_2017}. ConcpetNet provides commonsense knowledge. For example, if you query \texttt{<laptop, AtLocation, ?>}, it returns possible commonsense locations (\texttt{tail}), such as \texttt{desk}. As for the \texttt{head}, we used objects from Microsoft COCO objects~\cite{White2017UnifyingTS}, so that in the future we can also use object detection to collect visual data. 

In order to simplify the experiment setup, we decided to use a subset of the ConceptNet. To be more specific, we restricted the number of objects to 10, where the commonsense locations of every object is also restricted to be one. 10 random human names were used to mimic humans in the room. The relation $\bm{r}^{(t)}$ is always \texttt{AtLocation}. The four probabilities $p_{commonsense}$, $p_{new\_location}$, $p_{new\_object}$, and $p_{switch\_person}$ were set to $0.7$, $0.1$, $0.1$, and $0.5$, respectively. We have tuned these values to mimic a realistic environment. We have also set the maximum steps of the environment to $1,000$. This means that the environment terminates after an agent has taken $1,000$ steps.

\subsection{Single Agent Policies}
\label{sec:single_agent}

Inspired by the theories on the explicit human memory, we have designed the following four handcrafted policies (models).

\textbf{Handcrafted 1: Only episodic, forget the oldest and answer the latest}. This agent only has an episodic memory system. When the episodic memory system is full, it will forget the oldest episodic memory. When a question is asked and there are more than one relevant episodic memories found, it will use the latest relevant episodic memory to answer the question. 

\textbf{Handcrafted 2: Only semantic, forget the weakest and answer the strongest}. This agent only has a semantic memory system. When the semantic memory system is full, it will forget the weakest semantic memory. When a question is asked and there are more than one relevant semantic memories found, it will use the strongest relevant semantic memory to answer the question.

\textbf{Handcrafted 3: Both episodic and semantic}. This agent has both episodic and semantic memory systems. When the episodic memory system is full, it will forget similar episodic memories that can be compressed into one semantic memory. When the semantic memory system is full, it will forget the weakest semantic memory. When a question is asked, it will first try to use the latest episodic memory to answer it, if it can not, it will use the strongest relevant semantic memory to answer the question. 

\textbf{Handcrafted 4: Both episodic and pretrained semantic}. From the beginning of an episode, the semantic memory system is populated with the ConceptNet commonsense knowledge. When the episodic memory system is full, it will forget the oldest episodic memory. When a question is asked, it will first try to use the latest episodic memory to answer it, if it can not, it will use the strongest relevant semantic memory to answer the question.

For a fair comparison, every agent has the same total memory capacity. As for the Handcrafted 3 agent, the episodic and semantic memory systems have the same capacity, since this agent does not know which one is more important \textit{a priori}. As for the Handcrafted 4 agent, if there is space left in the semantic memory system after filling it up, it will give the rest of the space to the episodic memory system. In order to show the validity of our handcrafted agents, we compare them with the agents that forget and answer uniform-randomly.

\subsection{Multiple Agent Policies}
\label{sec:multiple_agent}

The multiple agent policies work in the same manner as the single agent policies, except that they can use their combined memory systems to answer questions.

\section{Results}
\label{sec:results}

\begin{figure}[tb]
    \centering
    \begin{subfigure}[b]{0.45\textwidth}
        \centering
        \includegraphics[width=\textwidth]{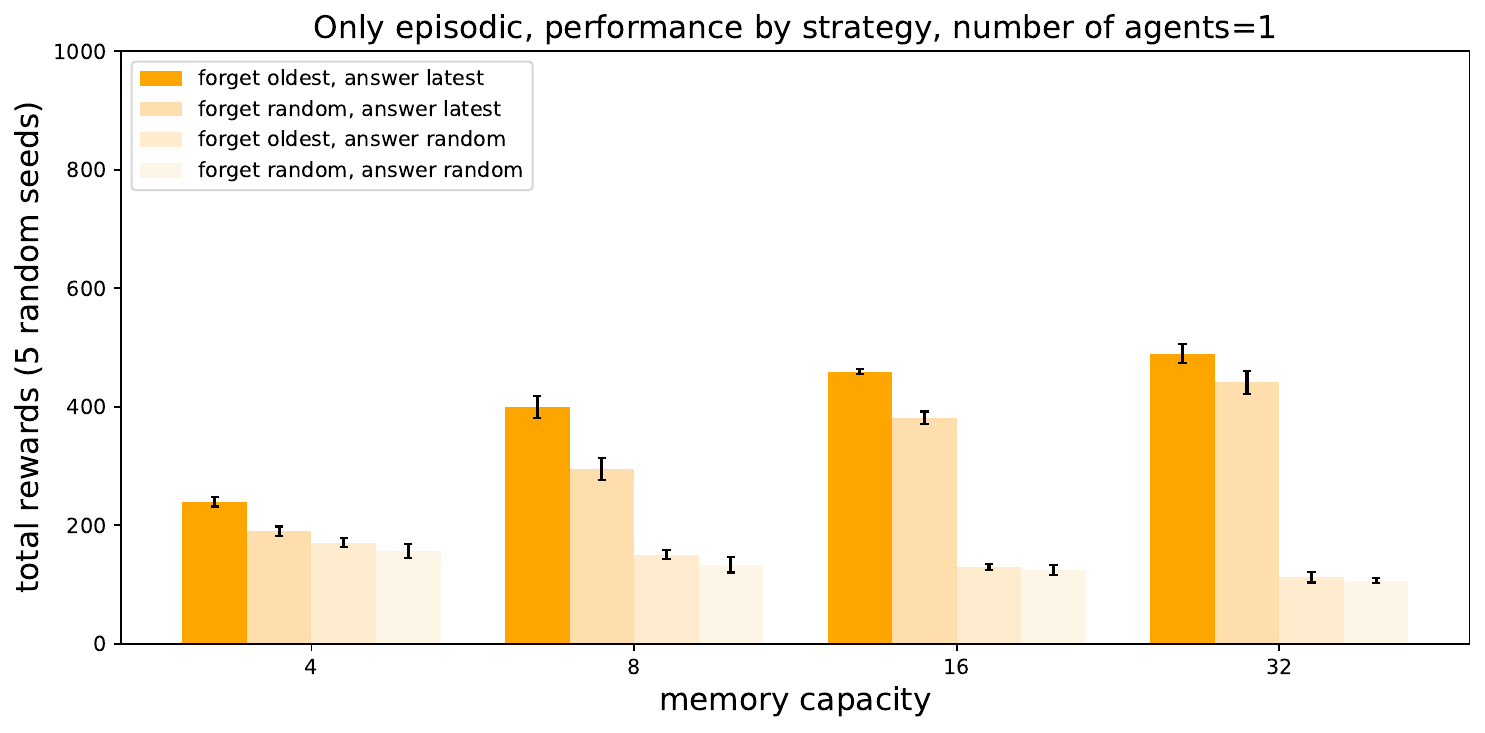}
        \caption{Handcraft 1}
        \label{fig:handcraft1_random}
    \end{subfigure}
    \hfill
    \begin{subfigure}[b]{0.45\textwidth}  
        \centering 
        \includegraphics[width=\textwidth]{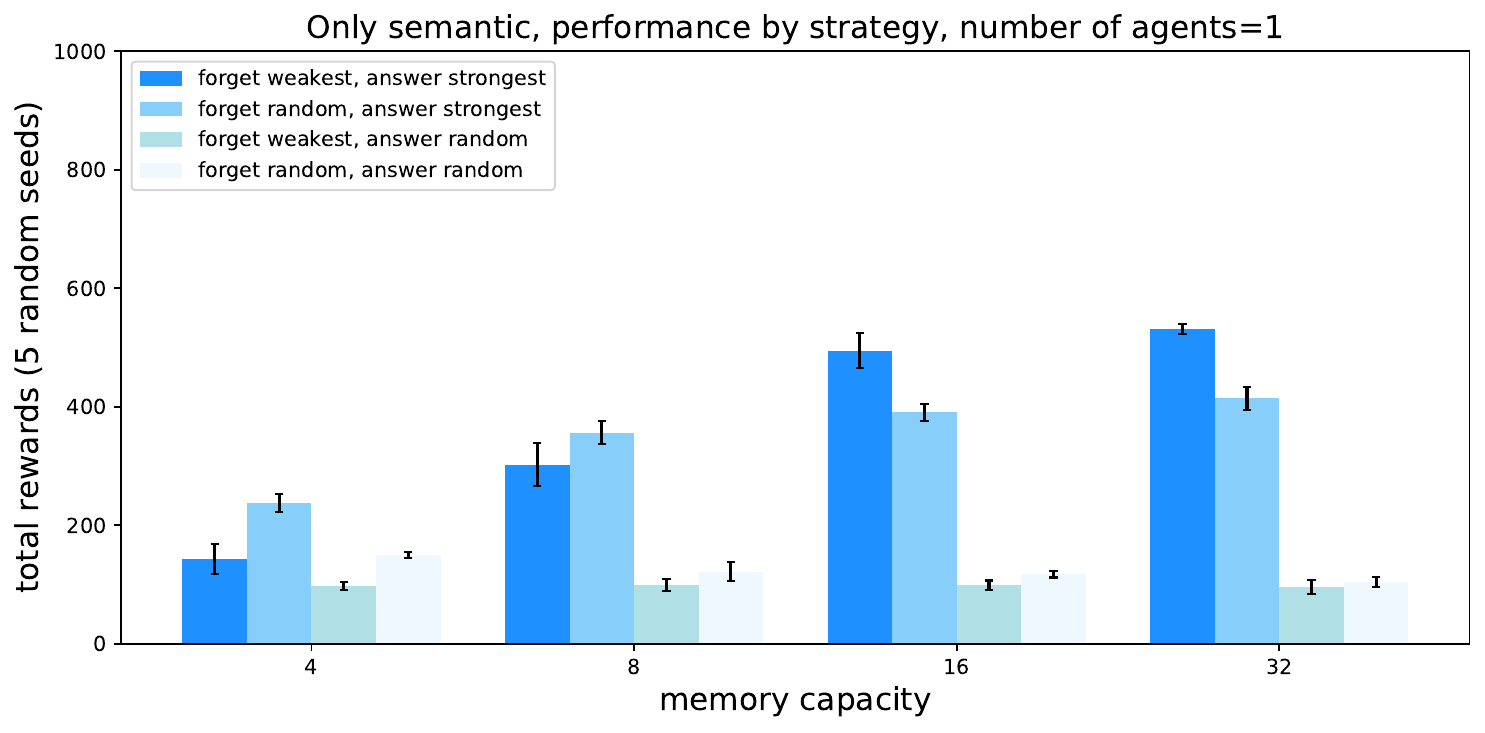}
        \caption{Handcraft 2}
        \label{fig:handcraft2_random}
    \end{subfigure}
    % \vskip\baselineskip
    \begin{subfigure}[b]{0.45\textwidth}   
        \centering 
        \includegraphics[width=\textwidth]{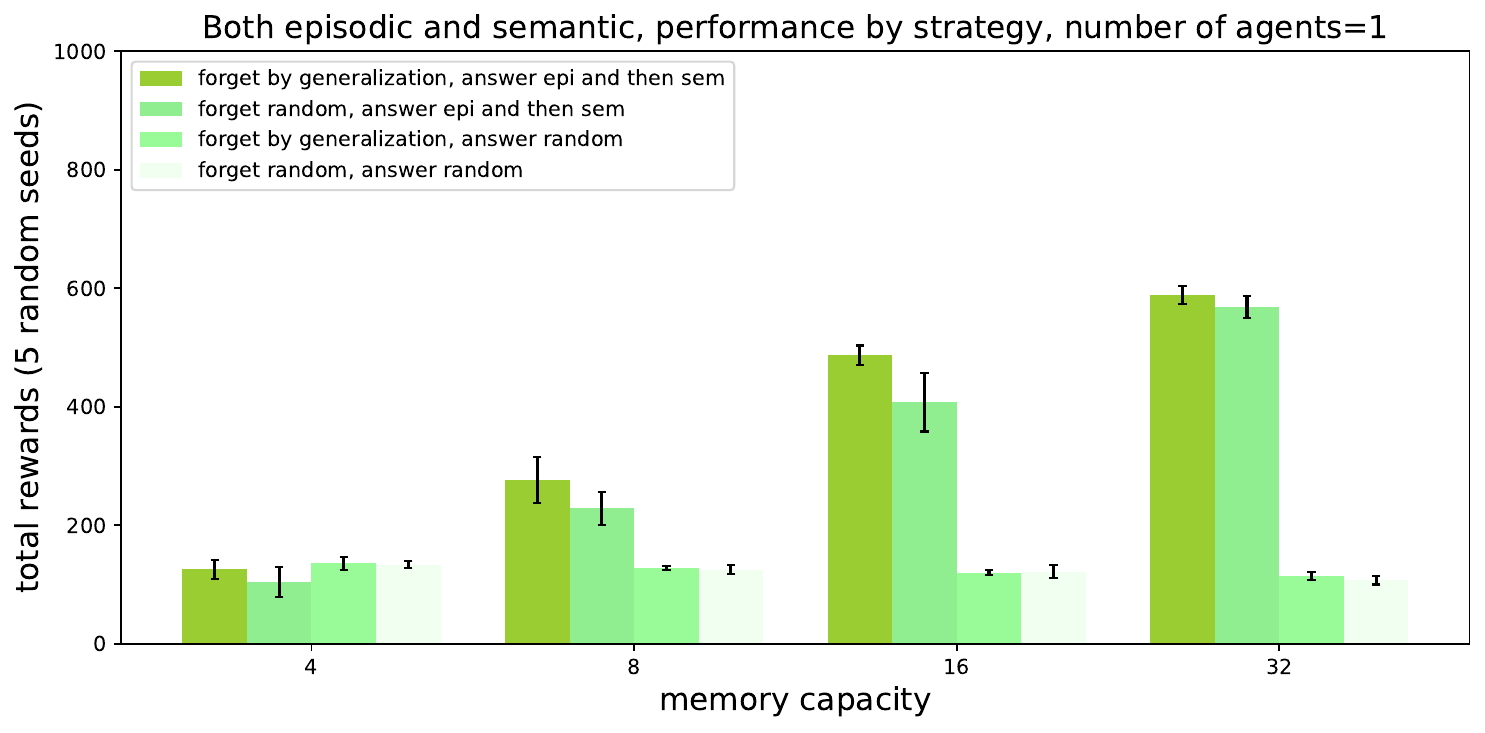}
        \caption{Handcraft 3}
        \label{fig:handcraft3_random}
    \end{subfigure}
    \hfill
    \begin{subfigure}[b]{0.45\textwidth}   
        \centering 
        \includegraphics[width=\textwidth]{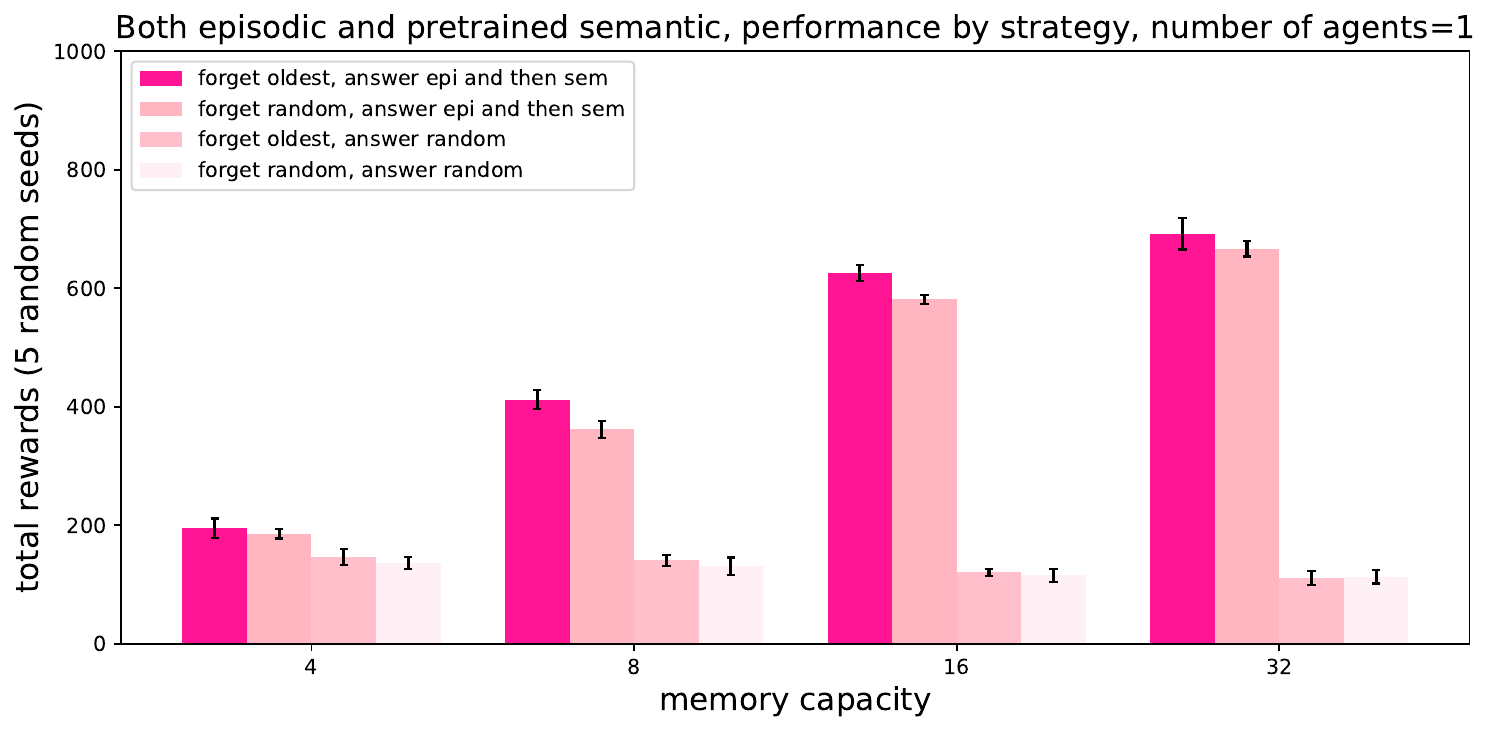}
        \caption{Handcraft 4}
        \label{fig:handcraft4_random}
    \end{subfigure}
    \caption{Handcrafted vs. random policies}
    \label{fig:handcrafted_random}
\end{figure}

Figure~\ref{fig:handcrafted_random} shows that our handcrafted forgetting and answering policies outperform random policies. Obviously, when both forgetting memories and answering questions are done randomly, it performs the worst. 

\begin{figure}[tb]
\centering
\includegraphics[width=\textwidth]{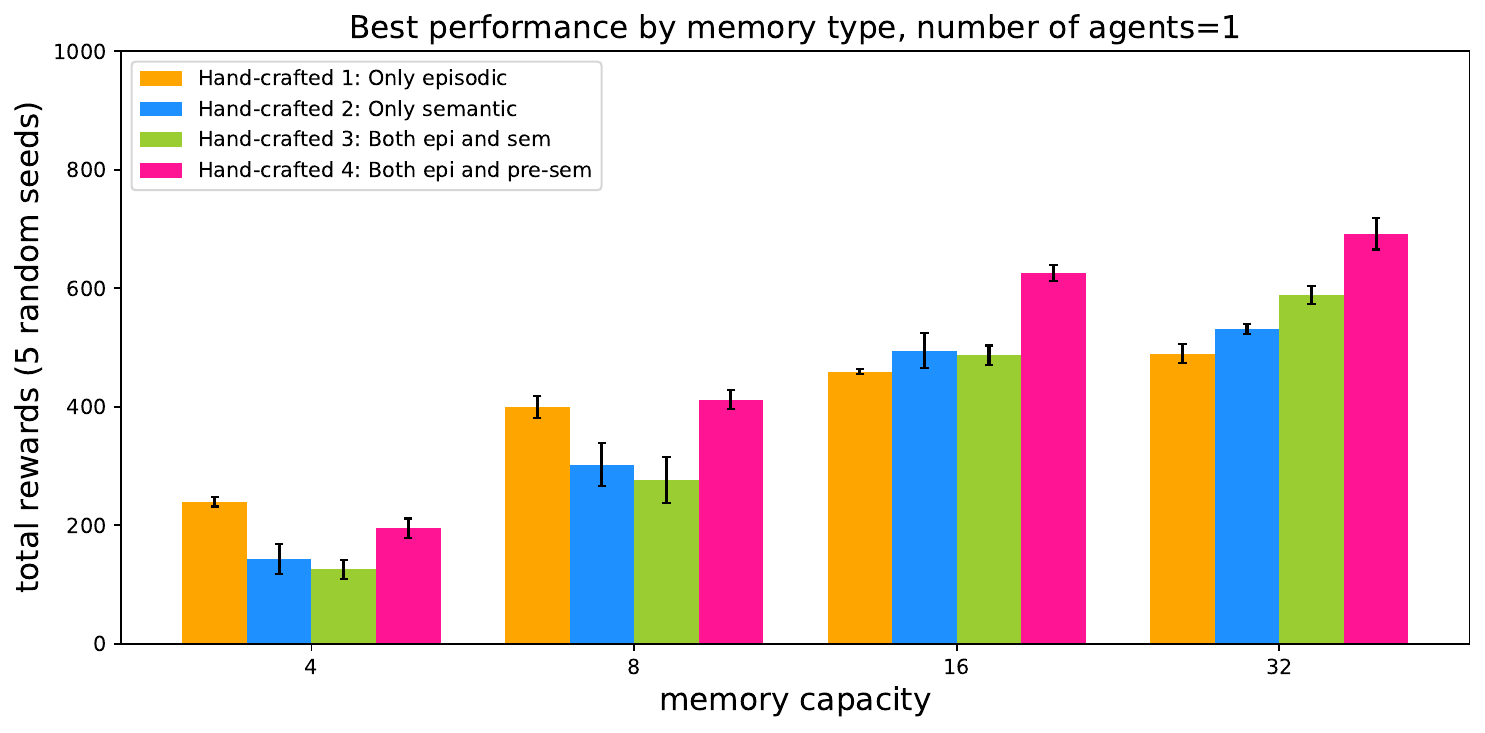}
\caption{Total rewards with respect to different handcrafted policies and memory capacities.}
\label{fig:best_strategies}
\end{figure}

Figure~\ref{fig:best_strategies} shows the results after one episode, with their best handcrafted policies. It shows that when the memory capacity is low, having only episodic memory system is better than the others. This is because when there are not enough memories in the system, it is not enough to learn the general world knowledge. As the memory capacity increases, however, it shows that having a semantic memory system helps, as it learns to generalize the world. It is especially interesting to see the Handcrafted 4 agent, which has an episodic system and a pretrained semantic system. Since it already knows the general world knowledge, it could focus more on the episodic part, leading to better performance.

\begin{figure}[tb]
\centering
\includegraphics[width=\textwidth]{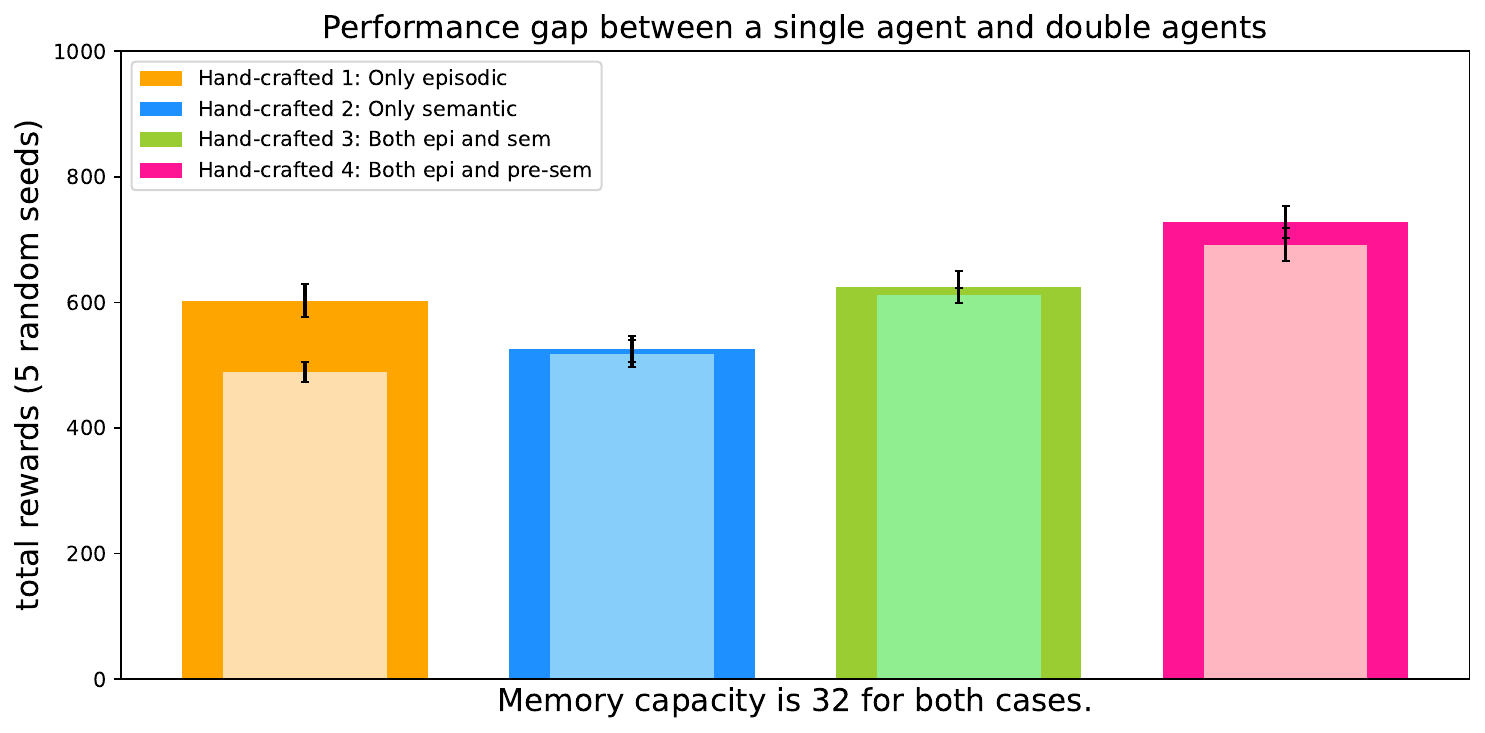}
\caption{Total rewards with respect to the number of agents. The lighter and narrower bars account for the single agent.}
\label{fig:single_and_double}
\end{figure}

Figure~\ref{fig:single_and_double} compares the performance between single-agent and double-agent setups. For a fair comparison, memory capacity was kept the same (i.e., as for the two agent setup, each agent can store 16 memories, while the agent in the single-agent setup can store 32). It shows that two agents working together were able to answer more questions than one agent working alone. This is due to the fact that the two agents were exploring the room in different directions. This led them to cover more area than one agent acting alone.

\section{Related work}
\label{sec:related}

After studying related literature, we observed that papers that are theoretically similar to ours are mostly cognitive science papers. ACT-R~\cite{ANDERSON2021101410} and Soar~\cite{10.5555/2222503} put a big emphasis on theories, but they lack of computational experiments, which makes it hard to compare. There was a work~\cite{Hemmer2009IntegratingEA} that studied how episodic and semantic memory systems play a role in recall of objects, but their experiments were human-based empirical results, which does not scale as well as our computational method.

Second is that although some computer science based papers do computational experiments that are a bit similar to ours, they often do not study episodic and semantic memory systems together. For example, Episodic Memory Reader~\cite{han-etal-2019-episodic} also learns what to forget in their memory system, and they also use question-answering to evaluate their method. However, this work only focuses on episodic memory. Also, their memory system is not composed of RDF-like data, but rather numeric embeddings, which are hard to interpret what they have captured.

\section{Conclusions}
\label{sec:conclusions}

We have created our own OpenAI-Gym-compatible environment, where agents with both episodic and semantic memory systems can be tested. We showed that when a machine is explicitly given both semantic and episodic memory systems, it outperforms the ones that only have one of the two memory systems. We also showed that when an agent is pretrained with commonsense knowledge, it outperforms the one that is not pretrained. Multiple agents collaborating with each other were better at question answering, since they can complement each other's memories.

In the future, we want to see if reinforcement learning agents can perform as good as the handcrafted ones, to see if such data-driven agents can lead to better generalization. We also hope to make the Room environment even closer to the real human environment (e.g., adding images, voices, more entities and relations, etc.). As for the collaboration, the current setup is limited to only agents working together. We would like to extend this to different collaboration setups (e.g., multiple humans and multiple agents). It would be especially interesting to encourage agents to ask humans questions, when it is not sure how to answer them. Collaboration will not always be straightforward, especially if there are conflicts and different degrees of trust among humans and agents. Dealing them elegantly will be another future challenge.

\section*{Acknowledgements}
\label{sec:ack}

This research was (partially) funded by the Hybrid Intelligence Center, a 10-year program
funded by the Dutch Ministry of Education, Culture and Science through the Netherlands Organization for Scientific Research, \href{https://www.hybrid-intelligence-centre.nl/}{https://www.hybrid-intelligence-centre.nl/}.

\bibliographystyle{vancouver}
\bibliography{bib}

\end{document}